\documentclass{ws-jmrr}
\usepackage{xcolor}
\usepackage[sort,compress,super]{cite}
\usepackage{url}
\usepackage{dirtytalk}
\begin{document}

\catchline{0}{0}{2013}{}{}

\markboth{Yamamoto et al.}{Tendon Actuated Concentric Tube Endonasal Robot (TACTER)}

\title{Tendon-Actuated Concentric Tube \\ Endonasal Robot (TACTER)}

\author{Kent K. Yamamoto$^{a}$\footnote{304 Research Drive, Durham, NC, 27708}\ , Tanner J. Zachem$^{a,b}$, Pejman Kheradmand$^c$, Patrick Zheng\textcolor{black}{$^a$},\\ \textcolor{black}{Jihad Abdelgadir$^b$}, Jared Laurance Bailey\textcolor{black}{$^a$}, Kaelyn Pieter\textcolor{black}{$^a$}, Patrick J. Codd$^{a,b}$, Yash Chitalia$^{c}$}

\address{$^a$Mechanical Engineering and Materials Science, Duke University, 304 Research Drive, \\ Durham, North Carolina, 27708, USA | E-mail: kky7@duke.edu}
\address{$^b$Department of Neurosurgery, Duke University School of Medicine, Durham, NC, USA}
\address{$^c$Speed School of Engineering, University of Louisville, Louisville, KY, USA}
\maketitle

\begin{abstract}
Endoscopic endonasal approaches (EEA) have become more prevalent for minimally invasive skull base and sinus surgeries. However, rigid scopes and tools significantly decrease the surgeon's ability to operate in tight anatomical spaces and avoid critical structures such as the internal carotid artery and cranial nerves. This paper proposes a novel tendon-actuated concentric tube endonasal robot (TACTER) design in which two tendon-actuated robots are concentric to each other, resulting in an outer and inner robot that can bend independently. The outer robot is a unidirectionally asymmetric notch (UAN) nickel-titanium robot, and the inner robot is a 3D-printed bidirectional robot, with a nickel-titanium bending member. In addition, the inner robot can translate axially within the outer robot, allowing the tool to traverse through structures while bending, thereby executing follow-the-leader motion. A Cosserat-rod based mechanical model is proposed that uses tendon tension of both tendon-actuated robots and the relative translation between the robots as inputs and predicts the TACTER tip position for varying input parameters. The model is validated with experiments, and a \textcolor{black}{human cadaver} experiment is presented to demonstrate maneuverability \textcolor{black}{from the nostril to the sphenoid sinus}. This work presents the first tendon-actuated concentric tube (TACT) dexterous robotic tool capable of performing follow-the-leader motion within natural nasal orifices to cover workspaces typically required for a successful EEA.
\end{abstract}


\begin{multicols}{2}

\section{Introduction} \label{sec:intro}
\subsection{Clinical Background}
The advent of endoscopic techniques have allowed for a new era of skull base and sinus surgery, specifically endoscopic endonasal approaches (EEA). By utilizing the natural passages of the nasal cavity and sinuses, regions of the para-nasal sinuses, both the anterior and middle skull base and cervico-occipital junction can be accessed with decreased morbidity as compared to their open counterparts [\citen{EEA}]. By utilizing neuro-navigation and high-definition endoscopes, surgical teams can approach these lesions, frequently resecting both benign and malignant neoplasms. Although EEA does not incur the risks associated with skin incisions and large craniotomies, a different set of complications is incurred, such as cerebrospinal fluid leaks, incomplete resection, and the requirements for complex skull base reconstruction, like tissue flaps. The most common EEA approach is the transsphenoidal resection of a Pituitary Adenoma. This typically benign tumor causes significant morbidity through its effect on hormonal function [\citen{EEA-JNS}] and cranial nerve dysfunction resulting from compression of critical nervous structures in the surrounding area. These tumors are located between the internal carotid arteries and multiple cranial nerves, making precision imperative, as aberrant movement can cause devastating injury to cranial nerves, vascular injury, stroke, and death. 

Additionally, incomplete resection can leave the patient with unresolved hormonal dysregulation, potentially requiring more invasive surgical interventions, and adjuvant therapies. As these tumors \textcolor{black}{grow} and invade the cavernous sinus and skull base, the rate of gross total resection (GTR) and endocrinological remission both decrease [\citen{knsop}]. One of the reasons for this is that tumor becomes non-linearly accessible, as the linear trajectories would endanger critical structures, such as the cavernous segment of the internal carotid artery and cranial nerves III, IV, V$_1$, V$_2$, and VI, as illustrated in Fig. \ref{fig:Intro}(a). Therefore, a method for safely accessing non-linearly approachable regions could make EEA safer, decrease the morbidity of treating these complex anatomical positions of tumor material, and increase the likelihood of GTR, endocrinological control, and a faster recovery. \textcolor{black}{The small workspace of the anterior skull base via endonasal approaches, combined with the high precision required due to the numerous critical neuroanatomical structures, as well as the size of the various pathologies, have created a strong clinical need for novel tools with increased flexibility and dexterity. New tools for accessing pathologies in nonlinear paths would allow surgeons to approach new, more advanced pathologies than currently possible.}
\vspace{0.4 cm}

\begin{figurehere}
\centering\includegraphics[scale = 0.59]{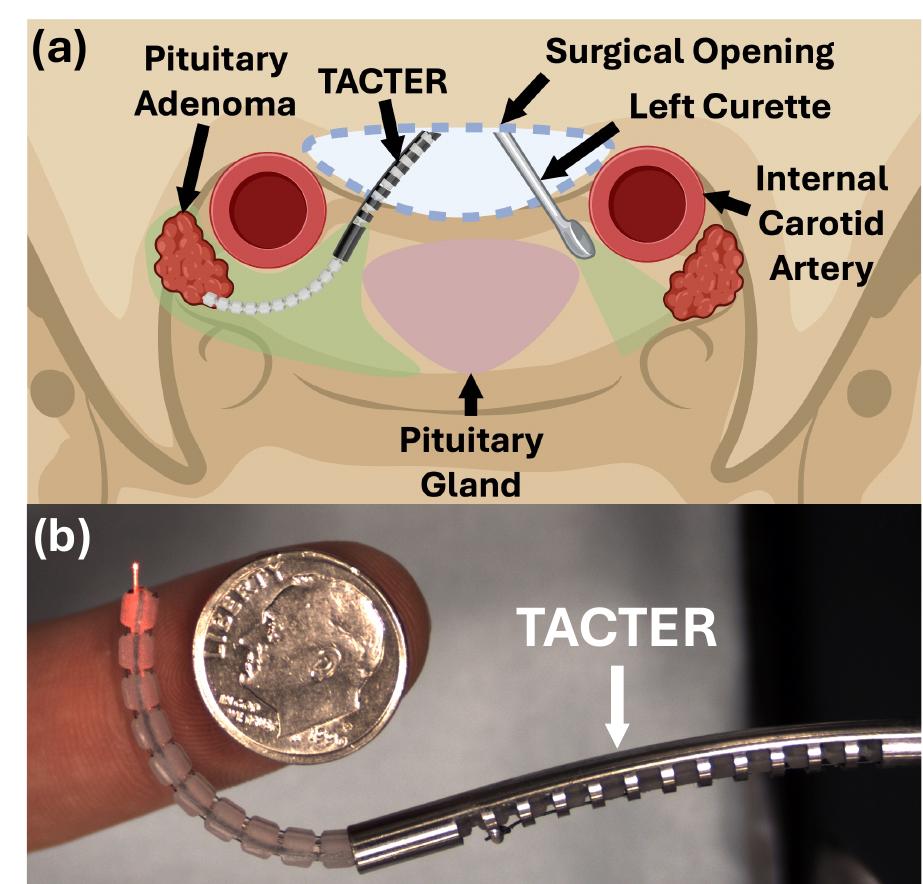}
    \caption{(a) Axial view of a pituitary adenoma resection, highlighting the advancements of the proposed steerable tool - the ability to access previously inaccessible regions compared to the current standard of care. (b) TACTER robot with both inner and outer joints actuated to produce an s-shaped curve around a U.S. dime. An optic fiber is threaded through the robot to demonstrate tooling capability at the tip.}
    \label{fig:Intro}
\end{figurehere}

\subsection{Continuum Robotics for Surgery}
Within novel minimally invasive surgical technologies research, continuum robotics theory has been a common foundation in designing steerable catheters, endoscopic tools, and needles [\citen{contMed1,contMed2,contMed3,CTER}]. Specifically, concentric tube robots (CTR) composed of pre-curved superelastic nickel-titanium alloy (nitinol, NiTi) tubes can provide surgeons the ability to navigate complex and anatomical trajectories inside the body. Another advantage of the CTR design is that follow-the-leader motion can be achieved due to each concentric tube having the ability to translate independently. However, snapping can occur due to multiple torques applied to each tube concentric to each other [\citen{snap1,snap2}], and there is limited dexterity once the robot reaches the task space for the surgery.

Another design of surgical continuum robotics is the tendon-actuated notched tube design [\citen{york2015wrist,TA2,TA3,TA4}]. Different geometry cutouts are made in a superelastic nitinol tube that allows compliance in different directions. By attaching \textcolor{black}{and pulling an actuation tendon at the tube tip,} a bending moment is applied, resulting in a windshield-wiping motion of the cut portion of the nitinol tube. \textcolor{black}{However, bending is limited to the notched portion, and curvilinear tip motion is lost as the moment is applied at a fixed point.} In summary, CTRs provide maneuverability to the surgical zone, and tendon-actuated notched robots increase dexterity at the tip.

There have been designs that incorporate both CTR and notched tube systems into a \textcolor{black}{\say{hybrid}} singular steerable robot. In [\citen{hybridLaser}], a traditional 3-tube CTR system with a tendon-actuated notched innermost tube allows for complex curvature with articulation at the end-effector. [\citen{COAST}] is a 3-tube steerable guidewire in which both the middle and outer tube is notched with an actuating tendon attached at the tip of the middle tube. A rigid steel inner tube adjusts how much of the notched region can bend, while the actuating tendon can then pull the middle tube to control the curvature of the robot. Finally, [\citen{TACTCatheter}] proposed attaching tendons to plastic catheters that are concentric for independent bending control of each catheter using tendon-actuation.

\textcolor{black}{Within endonasal approaches, a multitude of designs have been presented that incorporate both follow-the-leader capability and tip bending. [\citen{endoBC}] is a tendon-driven design with discrete spherical joints along the bending section (4 mm outer diameter, OD). However, it does not have follow-the-leader capability, as the bending section cannot translate axially. [\citen{endoFlex}] proposes a flat, pneumatic, soft robot design that has 5 degrees-of-freedom (DoF) for complex curvatures. However, widest portion of the robot is 10 mm, which would not be able to fit through the nostrils for an endonasal approach.}

\subsection{Tendon Actuated Concentric Tube (TACT) System}

This paper explores the first physical implementation of a multi-composite concentric tube robot (plastic and nitinol for the inner and outer robot respectively) in which both inner and outer robot bending can be independently controlled via tendon actuation, thus tendon-actuated concentric tube (TACT) system. \textcolor{black}{For endoscopic endonasal approaches, the robotic tool can be at most 4 mm in outer diameter to represent current endoscopic endonasal tools used [\citen{endoworld}], have follow-the-leader motion capability, and must have at least 2 DoF to curve around critical structures in the task space. The maneuverability possible with a TACT system can allow surgeons to snake around critical structures in the nasal cavity, such as the optic chiasm and cranial nerves in the cavernous sinus - hence tendon-actuated concentric tube endonasal robot (TACTER).} 

This paper will cover: The design of TACTER (Section \ref{sec:design}), a mechanical model previously proposed in [\citen{TACTModel}] to represent the end tip position of the inner robot given tendon tension of both inner and outer robots (Section \ref{sec:modeling}), experiments at twelve different unique poses to validate the mechanical model (Section \ref{sec: model validation}), and a \textcolor{black}{human cadaver} experiment to qualitatively demonstrate the workspace of the TACTER system (Section \ref{sec: cadaverExp}).

\section{Design} \label{sec:design}

\subsection{Robot Design and Fabrication}

\begin{figurehere}\label{fig:robotDesign}
\centering\includegraphics[scale = 0.6]{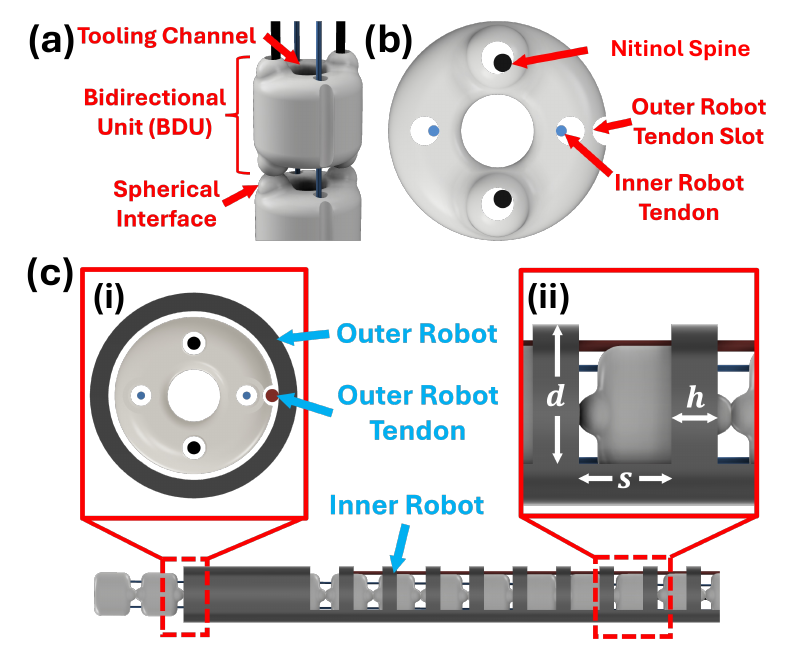}
    \caption{(a) 3D-printed inner robot design. (b) Inner robot routing and nitinol spine channels. c) TACTER design: i) outer robot routing schematic and ii) outer robot notch design parameters.}
\end{figurehere}

The tendon-actuated concentric tube endonasal robot (TACTER) comprises two independent tendon-actuated continuum robots, inner and outer, that are concentric to each other, as shown in Fig. \ref{fig:Intro}(b). The inner robot can translate about the outer robot and bend bidirectionally, and the outer robot can bend unidirectionally. \textcolor{black}{The proposed design currently allows both robots to bend about the same bending plane, and the user would rotate the robot for a full 360$^{\circ}$ workspace. The user would then manually operate both inner and outer robots to reach the desired target task space.}

%\subsubsection{Inner Robot}
The inner robot consists of 25 identical 3D-printed 3 mm outer diameter (OD) cylindrical segments (ProJet MJP 2500, 3D Systems Inc., Cary, NC, USA) with four holes 90$^{\circ}$ apart along the circumference with a 1.5 mm OD tooling channel, depicted in Fig. \ref{fig:robotDesign}(a). The two holes with spherical extrusions, similar to the design in [\citen{catheter_design}], are the routing holes for two 0.23 mm OD Nitinol rods that acts as spines, shown in Fig. \ref{fig:robotDesign}(b). The other two holes are to route 0.127 mm OD nitinol tendons for bidirectional actuation. A 0.2 mm radius semicircular slot is cut along the width of the cylinder planar to the two actuation routing holes. This cutout allows the routing of the outer robot actuation tendon without causing friction with the inner robot during translation. 

%\subsubsection{Outer Robot}
The outer robot is a tendon-actuated unidirectional asymmetric notched Nitinol joint, similar to the proximal joint in [\citen{pedNeuro}]. The inner robot is sheathed concentric to the outer robot, shown in Fig. \ref{fig:robotDesign}(c), and the actuation tendon in Fig. \ref{fig:robotDesign}(c-i) is soldered onto the last notch of the robot using flux for nitinol-soldering (Flux \#2, Indium Corporation, Oneida County, NY, USA). The notch design parameters - $d$, $h$, $s$ - shown in Fig. \ref{fig:robotDesign}(c-ii) are reported in Table \ref{table:geoParams}.

\subsection{Actuation System}
\textcolor{black}{
\begin{figurehere}
\centering\includegraphics[scale = 0.32]{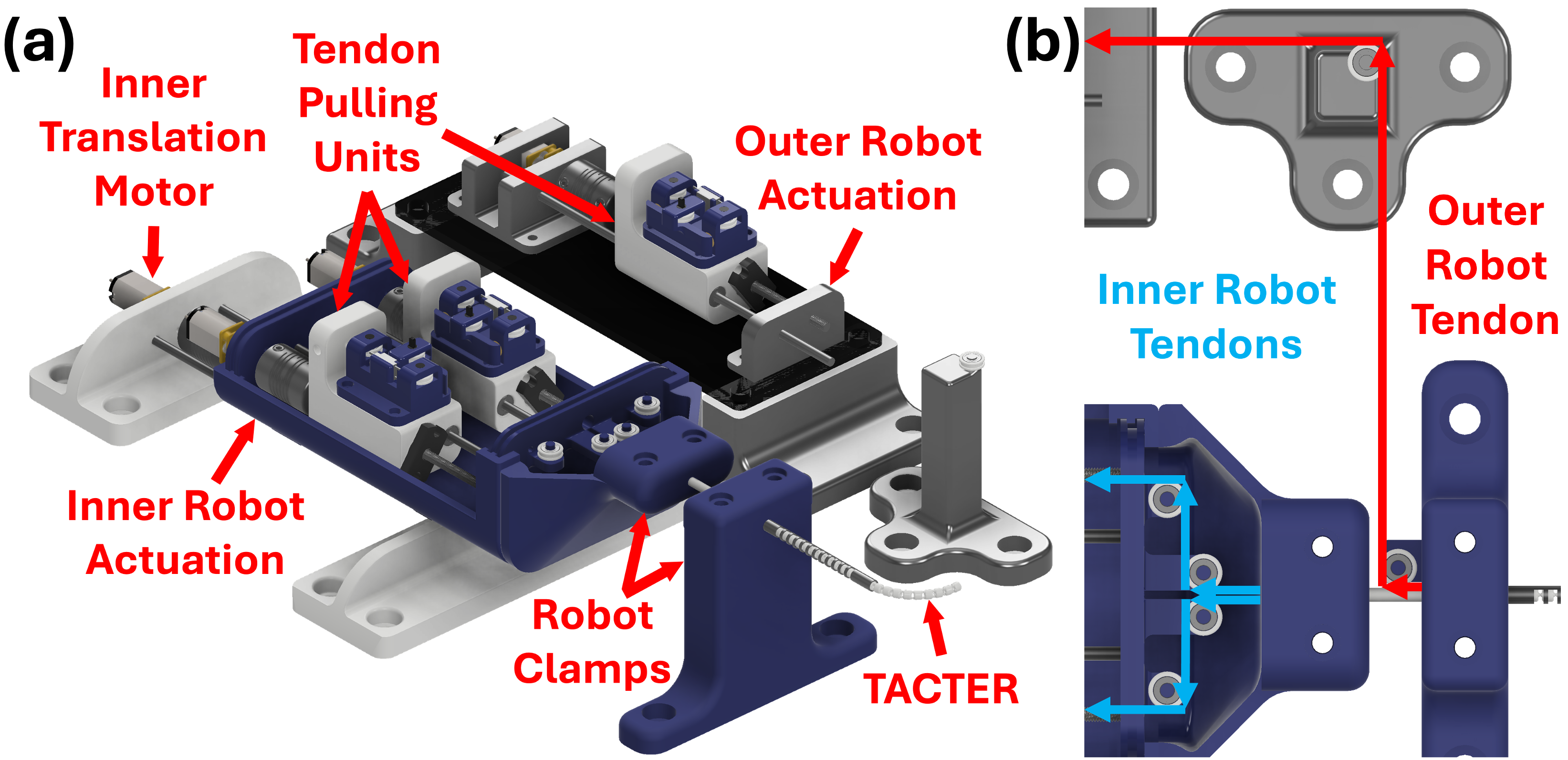}
    \caption{(a) TACTER benchtop actuation design rendering. (b) Inner and outer robot actuation routing.}
    \label{fig:actuation}
\end{figurehere}
}
The inner robot is actuated by pulling two nitinol tendons attached to the robot tip. Each tendon is pulled by a 3D-printed part that linearly translates on a lead screw and motor system, as shown in Fig. \ref{fig:actuation}(a). Micro-metal gear motors (298:1 gear ratio, Pololu Robotics and Electronics, Las Vegas, NV, USA) with magnetic quadrature encoders rotate a 0.6 mm pitch lead screw. The 3D-printed part has linear bearings and slides along a steel metal rod as it translates. A custom tension sensing unit is also attached to the two 3D-printed parts to measure the tendon tension for each actuating tendon (discussed in Section \ref{FS}). The inner robot actuation system is enclosed within a fixture that translates about the outer robot. The same micro-metal gear motor and lead screw system allow for up to 30 mm of inner robot translation. 

%\subsubsection{Outer Robot Actuation} 
The outer robot actuation tendon is routed along the notched side of the outer robot and through the outer grooves of the inner robot, leading to a separate outer robot actuation unit, illustrated in Fig. \ref{fig:actuation}(b). A pulley attached to the robot clamp reroutes the tendon directly out of the outer robot towards the actuation unit. 

\subsection{Tendon Tension Force Sensor} \label{FS}

To further improve the precision of the tendon-pulling actuation system, tendon tension sensing is implemented to record data to input to the proposed mechanical model. Inspired by [\citen{flexoTendon}], the design comprises two components, a frame and a pushing unit, with three pulleys. 

\begin{figurehere}
\centering\includegraphics[scale = 0.3]{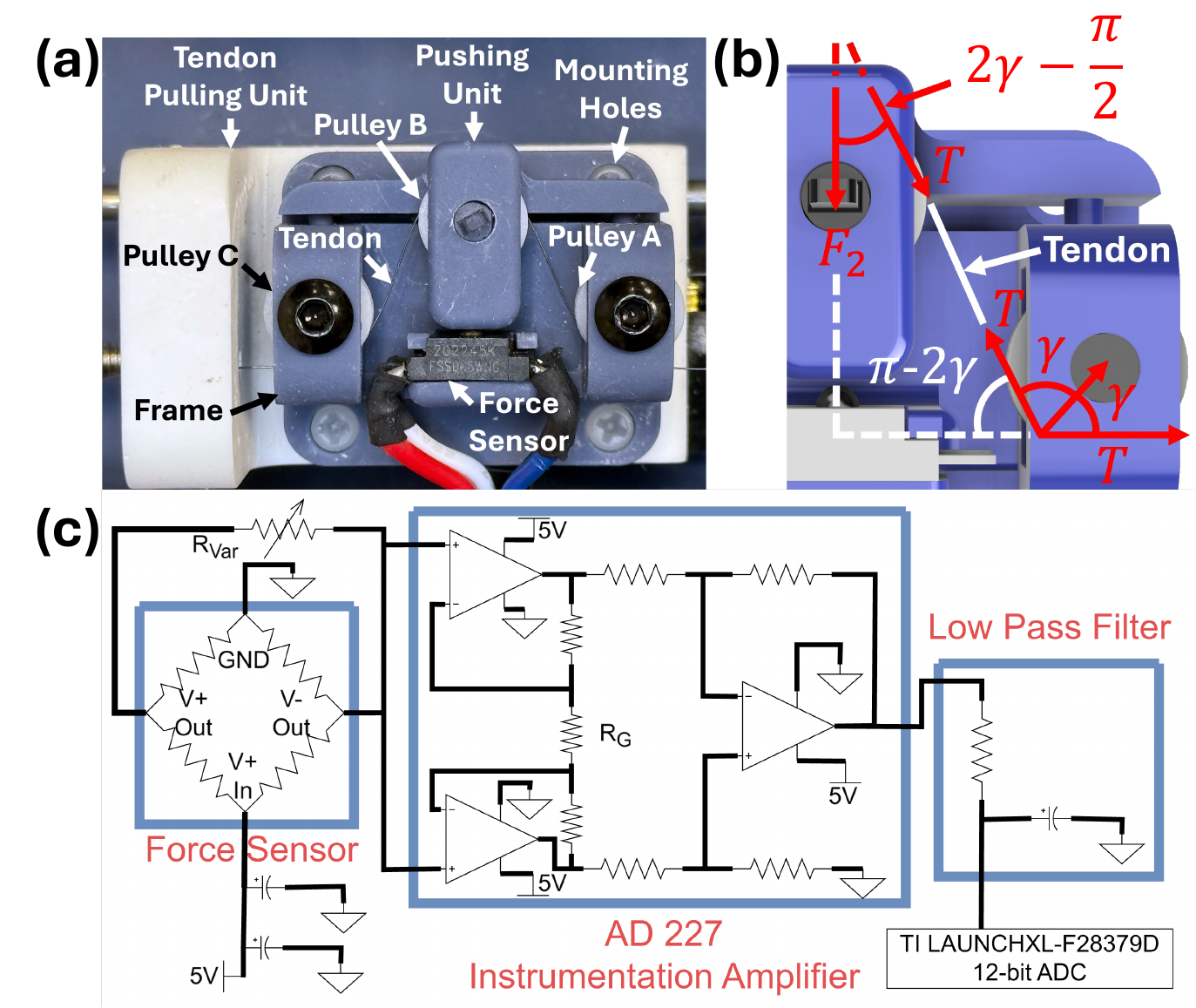}
    \caption{(a) Force sensor key components. (b) Geometric modeling to derive tendon tension in terms of force measured by the pushing unit. (c) Force sensor circuit schematic.}
    \label{fig:FS}
\end{figurehere}

The actuating tendon is first routed through the bottom pulley (A) of the frame, along the pushing unit pulley (B), and back down under the second pulley (C) of the frame, illustrated in Fig. \ref{fig:FS}(a). As the tendon is pulled, the pushing unit, guided by 3D-printed cylindrical extrusions in brass linear bearings, will translate linearly. The pushing unit applies a force down on a low-profile force sensor (FSS-SMT, Honeywell, Charlotte, NC, USA), resulting in a force value. Fig. \ref{fig:FS}(b) shows the geometric modeling done to convert the linear force read by the force sensor to tendon tension and results in:

\begin{equation}
    T_{t} = \frac{F_{B}}{2\sin(2\gamma)}
    \label{baseStroke}
\end{equation}
\noindent where $T_{t}$ is the tendon tension, $F_{B}$ is the force exerted on the force sensor due to the pushing unit, and $\gamma$ is the tendon routing angle from pulleys A and C to B ($\gamma = 52^{\circ}$).

%\subsubsection{Data Acquisition}
The circuit schematic, Fig. \ref{fig:FS}(c), illustrates the overall circuit design from sensing to processing. To stabilize the power supply and reduce voltage fluctuations, two parallel capacitors are placed across the supply line of the force sensor. The output signal from the force sensor is adjusted using a potentiometer to calibrate the baseline offset before amplification. This signal \textcolor{black}{inputs} into an instrumentation amplifier (AD8227, Analog Devices, Wilmington, MA, USA) for signal amplification with a gain of 13 ($R_G = 10k\Omega$). To minimize high-frequency noise, the amplified outputs are filtered using an RC low-pass filter (cutoff frequency of 20 Hz). The circuit is then redesigned onto a single printed circuit board (OSH Park, Lake Oswego, OR, USA), designed to support four force sensors.

\section{Mechanical Model} \label{sec:modeling}

\begin{figurehere}
\centering\includegraphics[scale = 0.5]{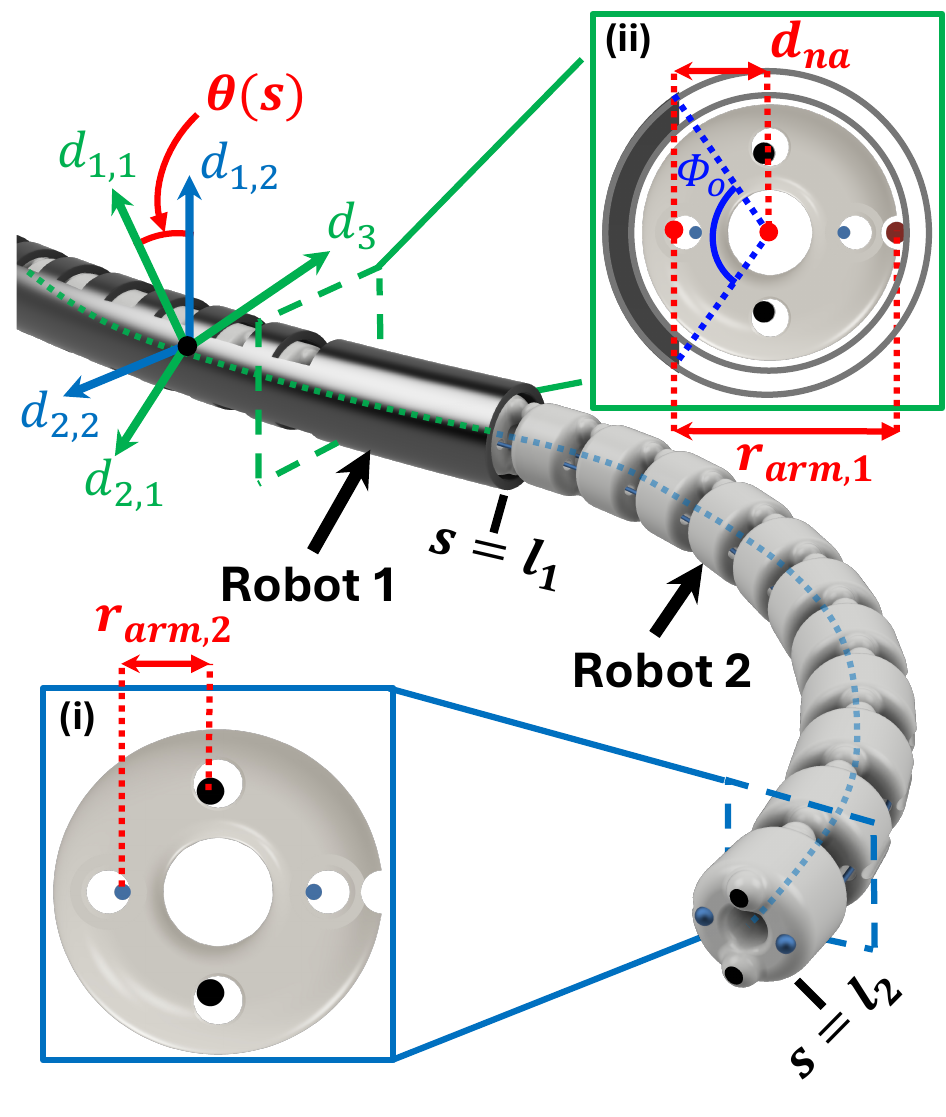}
    \caption{Coordinate frames and key design parameters for the proposed mechanical model.}
    \label{fig:model}
\end{figurehere}

TACTER comprises of two telescoping continuum robots (termed the outer and inner robots in Section \ref{sec:design}, above), each of which can be translated and rotated with respect to the other. In addition, individual robot bending is achieved by tensioning the tendons actuating each robot. Figure \ref{fig:model} shows a schematic of TACTER with the individual robot centerline frames. The arc-length shared by the two robots comprising TACTER is indicated by the variable $s$, such that the common base of both robots is at $s=0$. The outer robot terminates when $s=l_1$ and the tip of TACTER coincides with the tip of the inner robot, i.e., when $s=l_2$. The position of the individual robot centerlines is given by $P_i(s)$ and the material orientation of its director vectors $d_{1,i}, d_{2,i}$ and $d_3$ w.r.t an arbitrary global coordinate frame, is compactly represented by $R_i(s) \in \text{SO}(3)$. Individual robot position and orientations are a function of the translation, rotation and tendon tensions of the individual robots. For the portion of TACTER's length where the two robots overlap, we assume that they share a common centerline position $P(s)$. However, the individual tubes are allowed to twist with respect to each other, i.e., $R_2 = R_1R_{d_3}(\theta(s))$, where $\theta(s)$ is the twist angle between the outer and inner robots about their common director $d_3$. Similarly, the overall tangent vector to the robot centerlines ($\dot{P}(s)$) is also shared by both robots.
The spatial differential equations for the position and material orientation of the centerline of robot, $i$, (where $i=1$ for outer and $i=2$ for the inner robot) are given as follows:
\begin{align}\label{eq:shape_evol}
    \dot{P}(s) = R_i(s)v_i(s) \quad \dot{R}_i(s) = R_i(s)[u_i(s)]
\end{align}
Here, $u_i(s) \in \mathbb{R}^3$ and $v_i(s) \in \mathbb{R}^3$ are the angular and linear strain rates, as defined in [\citen{antman2005problems}]. Note, in the rest of the manuscript, as in Eq. (\ref{eq:shape_evol}), the dot notation indicates a spatial derivative, and $[ \ . \ ]$ is the skew-symmetric matrix representation of a cross product. 
 For a special Cosserat rod subjected to applied forces and moments per unit length given by $f \in \mathbb{R}^3$ and $\tau \in \mathbb{R}^3$ respectively, the equations for static equilibrium are indicated in [\citen{dupont2009design}] as follows:
\begin{align} \label{eq:equilibrium_eq_single_tube}
    \begin{bmatrix}
        \dot{n}_i \\
        \dot{m}_i
    \end{bmatrix}
    +
    \begin{bmatrix}
        [u_i] & 0 \\
        [v_i] & [u_i]
    \end{bmatrix}  
    \begin{bmatrix}
        n_i \\
        m_i
    \end{bmatrix} + 
    \begin{bmatrix}
        f_i \\
        \tau_i
    \end{bmatrix}=0
\end{align}
In this study, the distributed forces and moments are assumed to be only generated by the tendons. The position of each tendon in the cross-sectional body frame (indicated by the directors in Fig. \ref{fig:model}) is given by $r_{arm} = [x_{d1}, y_{d2}, 0]^T$. Assuming that these local body-frame coordinates stay constant along the arc-length $s$, the distributed force and moment \textcolor{black}{applied by the tendons} can be expressed as:
\begin{align} \label{eq:tendon_forces}
    f_{i} = - \lambda_{i}\frac{[\dot{p}_{i}^b]^2}{||\dot{p}_{i}^b||^3}\ddot{p}_{i}^b
\end{align} 
\begin{align} \label{eq:tendon_moment}
    \tau_{i} = - [r_{arm,i}]f_{i} = - [r_{arm,i}]\lambda_{i}\frac{[\dot{p}_{i}^b]^2}{||\dot{p}_{i}^b||^3}\ddot{p}_{i}^b
\end{align}
where $\lambda_{i}$ is the scalar tension in each tendon and $p_{i}^b$ represents the tendon position in the global frame. The derivatives of $p_{i}^b$ in the local frame is given by:
\begin{align} \label{eq:tendon_position_derivatives}
    \dot{p}_{i}^b = [u_i]r_{arm,i} + v_i, \quad
    \ddot{p}_{i}^b = [u_i]\dot{p}_{i}^b - [r_{arm,i}]\dot{u}_i + \dot{v}_i 
\end{align}
Since the two tubes are allowed to twist and elongate freely along $d_3$, their curvatures in the $d_1$ and $d_2$ directions must be equal. Additionally, $u_{d3}$, the third element of $u$, can be expressed in terms of the twist between the tubes,  $\dot{\theta}(s) = u_{2,d_3}(s) - u_{1, d_3}(s)$ and $\ddot{\theta}(s) = \dot{u}_{2,d_3} - \dot{u}_{1,d_3}$. Also $u_2$ can be expressed as:
\begin{align}\label{eq:u2_u1_relationship}
    u_2(s) = R^T_{d_3}(\theta)  u_1(s) + \dot{\theta}(s)\mathbb{I}_3
\end{align}
where $\mathbb{I}_3 = [0, 0, 1]^T$. The derivative of the above equation can be expressed as:
\begin{align} \label{eq:der_equal_curvature}
\dot{u}_2= (R^T_{d_3}(\theta) - \mathbb{I}_{(3,3)}) \dot{u}_1 + \dot{\theta} [\mathbb{I}_3]^T R^T_{d_3}(\theta) u_1 + \mathbb{I}_3 \dot{u}_{2,d_3}
\end{align}
Here, $\mathbb{I}_{(3,3)}$ represents a diagonal matrix where $\mathbb{I}_3$ forms its diagonal elements. By actuating the tendons, each tube undergoes compression, represented by a dilation factor. Following the convention of Gazzola et al. in [\citen{gazzola2018forward}], this factor is defined as $e_i = \frac{ds_i}{ds^*_i}$, where $ds$ is the element length in the deformed configuration and $ds^*$ is the corresponding length in the undeformed configuration. The relationship between the strains of tube 2 relative to tube 1 is then given by:
\begin{align}\label{eq:shear_relation}
v_2 = \beta R^T_{d_3}(\theta)v_1
\end{align}
Here $\beta$ is defined as the ratio $\beta = \frac{e_2}{e_1}$. Taking the derivative of the above equation, we obtain: 
\small
\begin{align}\label{eq:shear_dot_relation}
\dot{v}_2= \dot{\beta}R^T_{d_3}(\theta)v_1+\beta[\mathbb{I}_3]^TR^T_{d_3}(\theta)\dot{\theta}v_1+\beta R^T_{d_3}(\theta)\dot{v}_1
\end{align}
\normalsize

For concentric tubes, for static equilibrium, moments and forces applied by each tube to the other are additive in the $d_1$ and $d_2$ directions. With this assumption, the following equilibrium equations can be derived:
\begin{align} \label{eq:moment_sum_local1}
    \sum_{i = 1}^2 R_{d_3}(\theta)(\dot{m}_i+[u_i]m_i+[v_i]n_i+\tau_{i})=0|_{d_1,d_2}
\end{align}
and the sum of the forces in the local material frames is:
\begin{align} \label{eq:force_sum_local1}
    \sum_{i = 1}^2 R_{d_3}(\theta)(\dot{n}_i+[u_i]n_i+f_{i})=0|_{d_1,d_2}
\end{align}
Additionally, for each individual tube, the Eq. (\ref{eq:equilibrium_eq_single_tube}) can be applied in the $d_3$ direction.

\subsection{Constitutive Model}
TACTER can be modeled as a co-axially aligned set of two telescoping tendon-driven robots: 1) the outer robot, which is a unidirectional asymmetric notch joint robot, and 2) the 
inner robot, which is a 3D-printed set of cylindrical segments. For each robot, the following equations determine the linear relationships between shear-strain and shear-forces ($v_i$ to $n_i$), and bending-curvatures to bending-moments ($u_i$ to $m_i$):
\begin{align}\label{eq:constitutive_model_single}
     \begin{bmatrix}
        n_i \\
        m_i
    \end{bmatrix} = 
    \begin{bmatrix}
        K_{se,i} & 0 \\
        0 & K_{bt,i}
    \end{bmatrix}     
    \begin{bmatrix}
        v_i - v^*_i \\
        u_i - u^*_i
    \end{bmatrix}     
\end{align}
and,
\small
\begin{eqnarray} \label{eq:shear_const}
    K_{se,i} & = & 
    \begin{bmatrix}
        GA_i & 0 & 0 \\
        0 & GA_i & 0 \\
        0 & 0 & EA_i
    \end{bmatrix} ,   \\ \label{eq:moment_const}
    K_{bt,i} & = & 
    \begin{bmatrix}
        EI_{d1,i} & 0 & 0 \\
        0 & EI_{d2,i} & 0 \\
        0 & 0 & GJ_{d3,i} 
    \end{bmatrix}    
\end{eqnarray}
\normalsize
\textcolor{black}{,where the superscript '*' denotes the \say{undeformed} or initial value for each variable.}
The outer robot is a nitinol tube that is micromachined with a unidirectional asymmetric notch pattern (see Fig. \ref{fig:robotDesign}(ii)) [\citen{york2015wrist, chitalia2023model, chitalia2020design}]. A cross-section of the machined bending member is seen in Fig. \ref{fig:model}(inset (ii)). The micromachining depth $d$, the outer and inner radii ($r_{o,1}$ and $r_{i,1}$) determine the bending properties of this robot, namely $A_1$ and $I_1$. Electrical discharge machining (EDM) processes used to machine asymmetric notches into a cylindrical tube results in a machined region that is the difference between two circular segments of areas $ A_{out,1}$ and $ A_{in,1}$:
\begin{align}
    A_{out,1} &= \frac{r_{o,1}^2}{2}(\phi_o - \sin(\phi_o)) \nonumber \\
    A_{in,1} &= \frac{r_{i,1}^2}{2}(\phi_i - sin(\phi_i))  \\
    A_1 &= A_{out,1} - A_{in,1} \nonumber
\end{align}
Here, $\phi_o = 2cos^{-1}((d-r_{o,1})/r_{o,1})$ is the central angle subtended by the outer wall of the nitinol tube, and is explicitly shown in Fig. \ref{fig:model}(ii). Similarly, $\phi_i = 2cos^{-1}((d-r_{o,1})/r_{i,1})$ is the central angle subtended by the inner wall of the nitinol tube.
The neutral axis of this machined tube is located away from the centerline of the tube and is given by $d_{na}$:

%%%%% This needs fixing %%%%%

\begin{align}
d_{na} = & \frac{1}{A_{out,1} - A_{in,1}} \bigg( \frac{4A_{out,1}r_{o,1}\sin\left(\frac{\phi_o}{2}\right)^3}{3\left(\phi_o - \sin(\phi_o)\right)} \\ \nonumber 
& \quad - \frac{4A_{in,1}r_{i,1}\sin\left(\frac{\phi_i}{2}\right)^3}{3\left(\phi_i - \sin(\phi_i)\right)} \bigg)
\end{align}

Similarly, the second moment of area about the neutral axis is given as follows:
%%%%% This needs fixing %%%%%
\begin{align}
   I_{na} = \frac{r_{o,1}^4}{8}(\phi_o - \sin(\phi_o) + 2\sin(\phi_o)\sin(\frac{\phi_o}{2})^2) \\ \nonumber
    - \frac{r_{i,1}^4}{8}(\phi_i - \sin(\phi_i) + 2\sin(\phi_i)\sin(\frac{\phi_i}{2})^2)
\end{align}
\noindent and by the parallel axis theorem, the second moment of area about the robot centerline is given by $I_1 = I_{na} - A_1 d_{na}^2$.
Finally the moment arm of the outer robot's tendon to the neutral axis (see $r_{arm,1}$ in  Fig. \ref{fig:model}(ii)) is given as follows:
\begin{align}
    r_{arm,1} = d_{na} + r_{i,1} - r_{tendon, 1}
\end{align}
where $r_{tendon, 1} = 80$~$\mu$m is the radius of the actuating tendon of the outer robot.

The inner robot is actuated by two tendons passed through channels printed within the cylindrical segments that form the body of the robot. However, flexion of this inner robot is achieved through the two nitinol rods that are routed in a plane orthogonal to the plane of the tendons. Therefore, only the cross-section of these nitinol rods must be taken into consideration while computing cross-sectional area ($A_2$) and second moment of area ($I_2$) for this robot. These are given as follows:
\begin{align}
A_2 = 2\pi r_{rod}^2, \quad
I_2 = \frac{\pi r_{rod}^4}{2}
\end{align}
where, $r_{rod} = 115$~$\mu$m is the radius of the nitinol rods that act as the inner robot's bending member. Similarly, the moment arm of the inner robot's tendons is simply the distance between the tendons and these rods, when the tendon is actuated and the tendons run along the walls of the channel closest to the spines. As a result, this moment arm, seen in Fig. \ref{fig:model}(i) is given by $r_{arm,2} = 0.725$~mm.
Note, that bending behavior of both robots comprising the TACTER is anisotropic, i.e. $I_{d1, i} \neq I_{d2, i}$ for both the outer and inner robots (i.e. $i=1, 2$). However, for the purposes of this paper, since only in-plane bending experiments are considered, isotropy has been assumed.

\begin{tablehere}
\centering
\tbl{Mechanical Model Geometric Parameters}
{\resizebox{\columnwidth}{!}{
\begin{tabular}{|c||c|c|c|}
\hline
Variable & Description  & Outer & Inner \\
\hline
\hline
$d$ & Notch Depth & 2.93 mm  & --\\
\hline
$h$ & Notch Spacing & 0.96 mm & -- \\
\hline
$s$ & Notch Width & 1.96 mm & -- \\
\hline
$r_{rod}$ & Nitinol Spine Radius & -- & 0.115 mm \\
\hline
$r_{0}$ & Outer Radius & 1.97 mm & 1.51 mm\\ 
\hline
$r_{i}$ & Inner Radius & 1.60 mm & 0.45 mm \\
\hline
$r_{arm}$ & Moment Arm &  3.06 mm & 0.725 mm\\
\hline
$r_{tendon}$ & Tendon Radius &  0.08 mm & 0.002 mm \\
\hline
$E$ & Elastic Modulus & 84 GPa & 100 GPa \\
\hline
$G$ & Shear Modulus & 28.8 GPa & 28.8 GPa \\
\hline
\end{tabular}}}
\label{table:geoParams}
\end{tablehere}

\subsection{Numerical Solution}
Substituting Eqs.(\ref{eq:tendon_forces}), (\ref{eq:tendon_moment}), (\ref{eq:u2_u1_relationship}), (\ref{eq:der_equal_curvature}), (\ref{eq:shear_relation}) and (\ref{eq:shear_dot_relation}) into Eqs. (\ref{eq:moment_sum_local1}) and (\ref{eq:force_sum_local1}), along with the individual equations derived from Eq. (\ref{eq:equilibrium_eq_single_tube}) in the $d_3$ direction, yields the following form:
\begin{align} \label{eq:finall}
\begin{bmatrix}
\mathbb{G}_u & \mathbb{G}_v & \mathbb{G}_{u_{d_3,2}} & \mathbb{G}_{\beta}\\
\mathbb{H}_u & \mathbb{H}_v & \mathbb{H}_{u_{d_3,2}} & \mathbb{H}_{\beta}\\
\mathbb{J}_u & \mathbb{J}_v & \mathbb{J}_{u_{d_3,2}} & \mathbb{J}_{\beta}\\
\mathbb{K}_u & \mathbb{K}_v & \mathbb{K}_{u_{d_3,2}} & \mathbb{K}_{\beta}
\end{bmatrix}
\begin{bmatrix} 
\dot{u}_1 \\ \dot{v}_1 \\ \dot{u}_{d_3,2} \\ \dot{\beta} 
\end{bmatrix}=
\begin{bmatrix} 
RHS_1 \\  RHS_2 \\ RHS_3 \\ RHS_4  
\end{bmatrix}
\end{align}
where all $\mathbb{G}$, $\mathbb{H}$, $\mathbb{J}$, $\mathbb{K}$ and $RHS$ are functions of the state variables [\citen{TACTModel}]. To solve the differential Eq. (\ref{eq:finall}), boundary conditions, $u_1(0)$, $v_1(0)$, $u_{d_3,2}(0)$ and $\beta(0)$ are required. Since these values are unknown, a shooting method is employed. In this approach, initial guesses for these values are made, and the equations are then solved using the kinematic inputs $\lambda_{1}$, $\lambda_{2}$ and known boundary conditions $R(0)$, $P(0)$ and $\theta(0)$.
Tendon termination at $s = l_i$, exerts a point force and moment given by:
\begin{eqnarray} \label{eq:boundry_tendon_forces}
    f_{i}(l_i) = - \lambda_{i}\frac{\dot{p}_{i}^b(l_i)}{||\dot{p}_{i}^b(l_i)||}\\
    \tau_{i}(l_i) = - [r_{arm,i}]\lambda_{i}\frac{\dot{p}_{i}^b(l_i)}{||\dot{p}_{i}^b(l_i)||}
    \label{eq:boundry_tendon_moment}
\end{eqnarray}
The boundary condition at the endpoint is specified by:
\begin{align}\label{eq:boundary_n}
 n_i(l^-_i) = n_i(l^+_i) + f_{i}(l_i)
\end{align}
\begin{align}\label{eq:boundary_m}
 m_i(l^-_i) = m_i(l^+_i) + \tau_{i}(l_i)
\end{align}
Here, $l^{\mp}_i$ represents the arc-length values infinitesimally before and after $l_i$ [\citen{D_Caleb_Rucker}]. Any guess that satisfies Eqs. (\ref{eq:boundary_n}) and (\ref{eq:boundary_m}) at the endpoint can be considered the correct guess for Eq. (\ref{eq:finall}), which can then be solved numerically.

\section{Experiments and Results} \label{sec:results}

\subsection{Experiment Design}
\textcolor{black}{Thirteen configurations} are tested to validate the mechanical model: a combination of outer robot straight/bent (OS/OB), inner robot fully sheathed/half translated/fully translated (IN/IH/IF - \textcolor{black}{0 mm/15.90 mm/30.22 mm for OS; 0 mm/15.28 mm/30.36 mm for OB}), and inner left/right actuation (L/R). For example, OB-IF-R will refer to the TACTER outer robot bent, inner robot fully translated, and the right inner robot tendon \textcolor{black}{loaded}. \textcolor{black}{The inner robot without the outer robot is also considered in the thirteen \textcolor{black}{configurations}.} The inner robot is translated to each respective scenario before actuation and data acquisition. \textcolor{black}{Both robots are actuated to their maximum tendon stroke (4 mm for inner, 7 mm for outer robot) in 15 steps with a tendon stroke of 0.267 mm and 0.467 mm per step, respectively.} Motor position, tendon stroke, and tendon tension are measured and recorded in Simulink for each scenario.

\vspace{0.3cm}
 \textcolor{black}{
\begin{figurehere}
\centering\includegraphics[scale = 0.45]{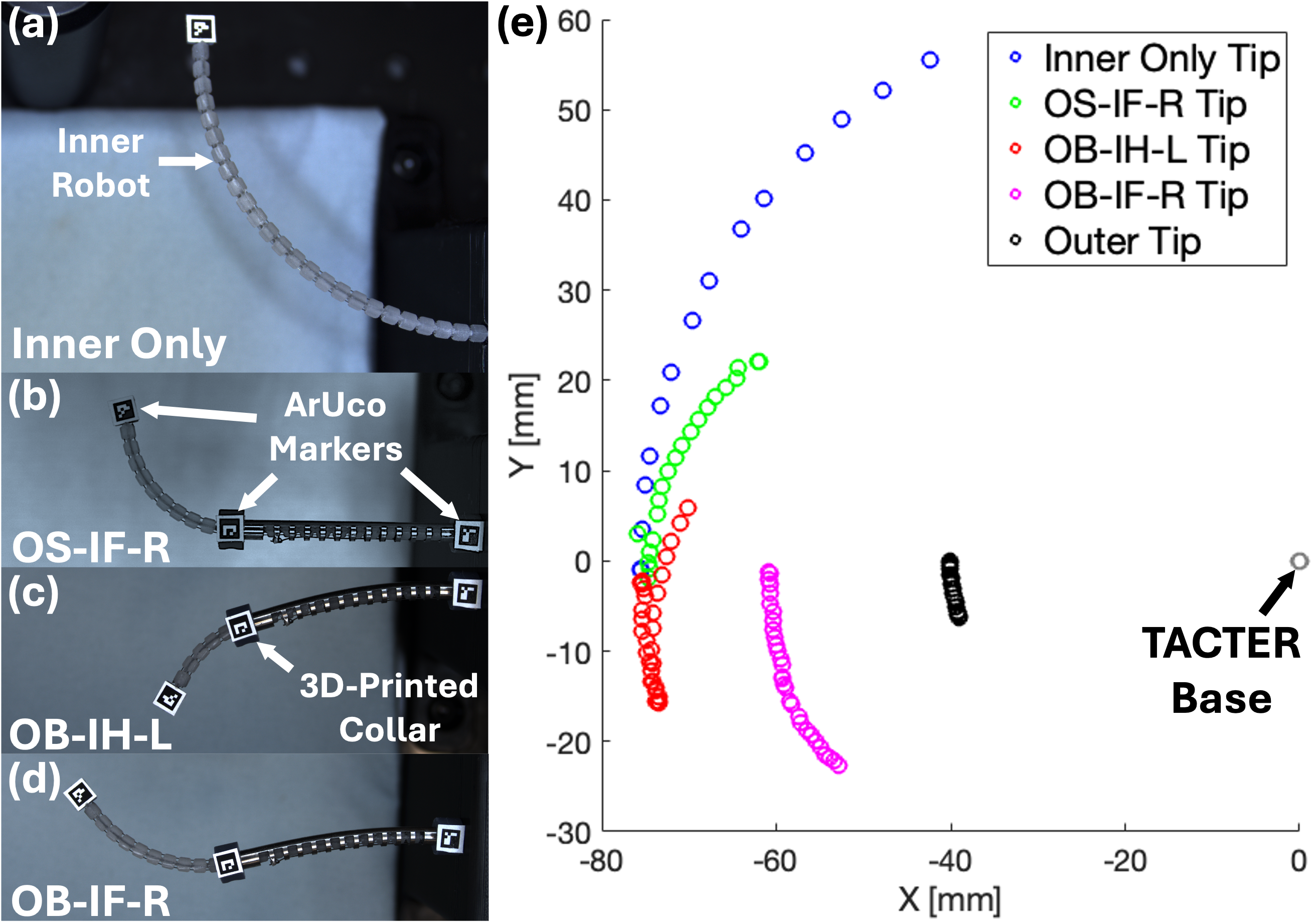}
    \caption{a) Fully actuated images of inner robot only, (b) Outer Straight, Inner Full, Right (OS-IF-L),(c) Outer Bent, Inner Half, Left (OB-IH-L), and (d) Outer Bent, Inner Full, Right (OB-IF-R). (e) 2D comparison plot of tip position during tendon loading for the \textcolor{black}{configurations} mentioned in (a-d).}
    \label{fig:exp}
\end{figurehere}
}
A camera (1800 U-2050C, Allied Vision Technologies GmbH, Stadtroda, Germany) is used to capture images of the robot at each actuated pose. 3D-printed collars with AruCo markers, shown in \textcolor{black}{Fig. \ref{fig:exp}(b) and \ref{fig:exp}(c)}, are attached at the base and tip of the outer robot to obtain the position of said points. Additionally, an end-effector attachment with an AruCo marker that fits inside the tooling channel of the inner robot is attached to obtain the inner robot tip position. Both position offsets from the collar and end-effector attachments are implemented post-data acquisition.

\textcolor{black}{A 2D plot of the tip position during loading for four of the thirteen \textcolor{black}{configurations} is presented in Fig. \ref{fig:exp}(e). The plot shows the ability for the outer robot to bend, inner robot to translate out, then bend bidirectionally. Although the bending of the inner robot only has a greater bending angle, the lack of an outer robot and ability to translate reduces the inner robot's ability to traverse in curvilinear trajectories.}

\subsection{Mechanical Model Validation} \label{sec: model validation}

\begin{figurehere}
\centering\includegraphics[scale = 0.36]{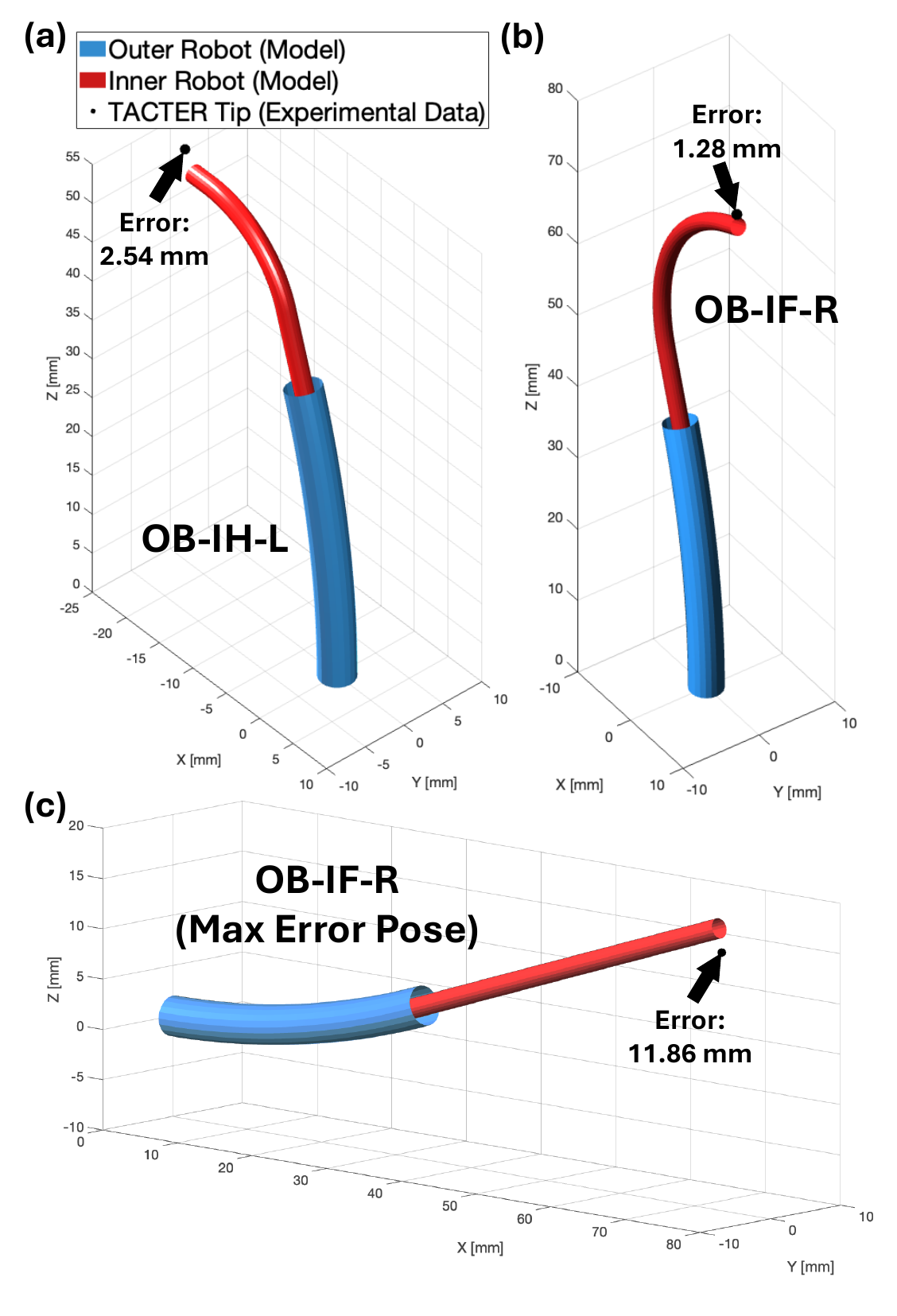}
    \caption{TACTER comparison between model tip prediction and experimental results for (a) OB-IH-L and (b) \textcolor{black}{OB-IF-R at their fully actuated states. (c) The pose with the greatest error among all trials, found during the loading phase of the OB-IF-R configuration (during actuation of the inner robot at 5.137 mm of tendon displacement.}}
    \label{fig:results}
\end{figurehere}

\textcolor{black}{
\begin{tablehere}
\tbl{TACTER Mechanical Model Validation Results \label{table:expResults}}
{\begin{tabular}{ |c||c|c|c||c|c|c|  }
 \hline
   & \multicolumn{3}{|c||}{Outer Straight} & \multicolumn{3}{|c|}{Outer Bent} \\
 \hline
 Error (mm) & IN & IH & IF & IN & IH & IF \\
 \hline
 Avg. & 0.75  & 2.14  & 4.05 & 0.69  & 1.33  & 2.38 \\
 \hline
 Max & 1.30 & 4.59  & 10.53 & 1.68 & 4.04 & 11.86 \\
 \hline
\end{tabular}}
\end{tablehere}
}

The Euclidean distance between measured TACTER tip position and model tip position is calculated at 30 discretized loading poses for each of the \textcolor{black}{thirteen configurations}. The left and right actuation trials are grouped together and averaged. Both average and maximum Euclidean distance error for OS and OB are reported in Table \ref{table:expResults}. Fig. \ref{fig:results} shows the full actuation model spline as a 3D rendering and the measured tip position for two specific \textcolor{black}{configurations}: OB-IH-L (Fig. \ref{fig:results}(a)) and OB-IF-R (Fig. \ref{fig:results}(b)). Fig. \ref{fig:results}(c) is the position with the greatest error out of the experiments (from the OB-IF-R scenario).

The results show that the average error between model and physical robot is relatively low (approximately 2.5\% of robot length) when the inner robot is fully sheathed. As the inner robot translates further out of the robot, there is a noticeable increase in error, especially when calculating the max error within the 30 poses (under 15\% of robot length). However, the average error stays less than 5 mm for all OS/OB-IN/IH/IF configurations.

\subsection{\textcolor{black}{Human Cadaver Experiment} \label{sec: cadaverExp}}

\textcolor{black}{A human cadaver experiment is conducted to assess clinical feasibility. Both the robot and a 5 mm OD, 0$^{\circ}$  endoscope are inserted into the right nostril of the cadaver model, shown in Fig. \ref{fig:cadaver}(a). The outer robot is first actuated to adjust the angle of attack  for the inner robot (Fig. \ref{fig:cadaver}(b)), then the inner robot is translated outwards to traverse medial to the middle turbinate towards the sphenoid ostium, captured in Fig. \ref{fig:cadaver}(c). Finally, the inner robot is actuated to bend to the right into the sphenoid ostium, as seen in Fig. \ref{fig:cadaver}(d).}
 
To demonstrate implementation of spectral tissue sensing such as [\citen{TumorID-SR,TumorID-Plos,TIDTucker}], a laser fiber (SM600, ThorLabs, Newton, NJ, USA) with a 635~nm 0.90~mW laser (PL202, ThorLabs, Newton, NJ, USA) coupled via a 635~nm Air-Spaced Doublet Collimator (F810APC-635, ThorLabs, Newton, NJ) is fed through the tooling channel of the inner robot. \textcolor{black}{The inner robot with laser fiber is shown in Fig. \ref{fig:cadaver}(e).}

\textcolor{black}{
\begin{figurehere}
\centering\includegraphics[scale = 0.43]{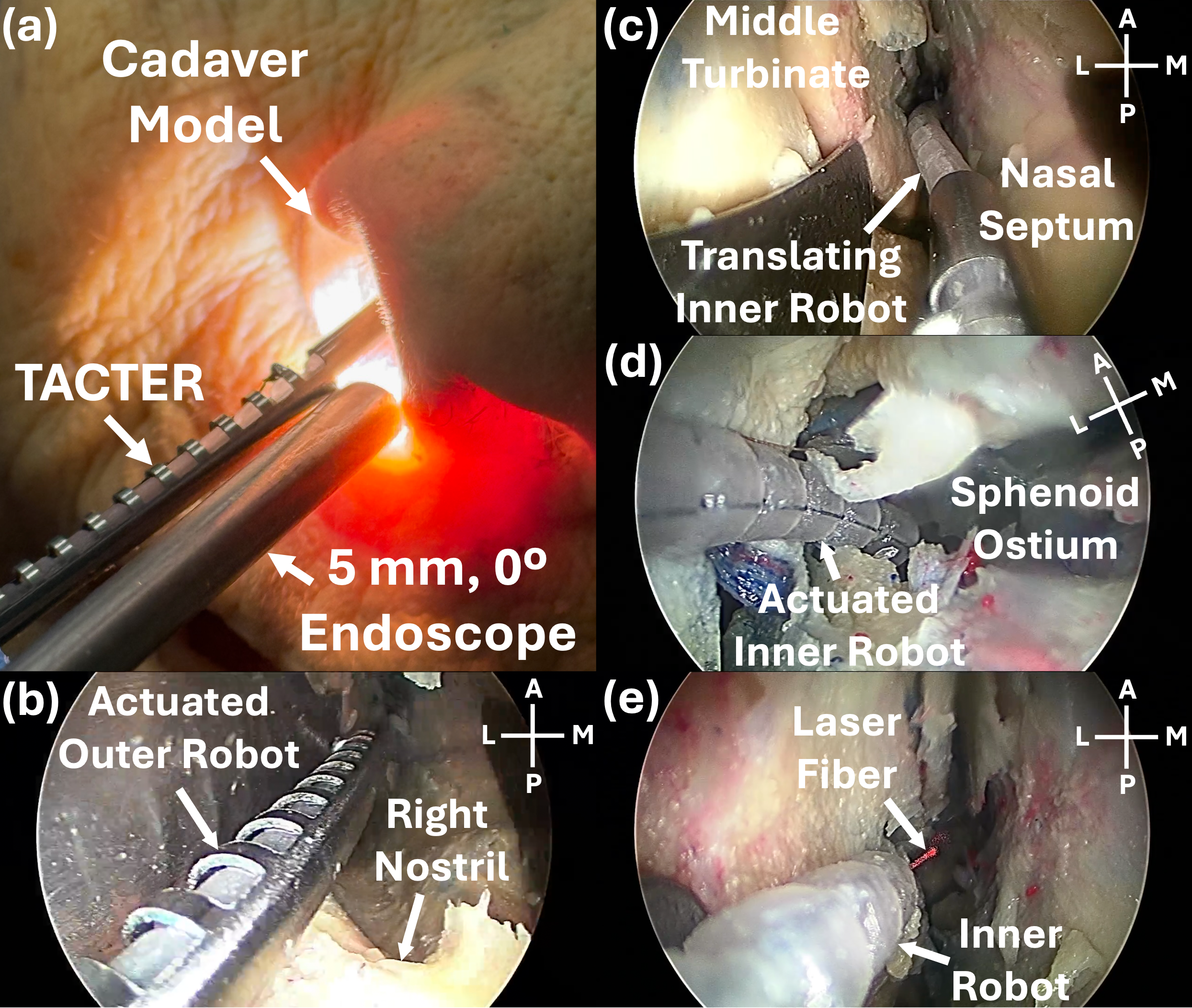}
    \caption{(a) TACTER and a 5 mm, 0$^{\circ}$ endoscope entering the right nostril. (b) Outer robot actuated inside the nasal passageway. (c) The inner robot translated out and traversing between the middle turbinate and nasal septum. (d) The inner robot actuated and entering the sphenoid sinus. (e) Demonstration of integrating an optical fiber through TACTER for future work.}
    \label{fig:cadaver}
\end{figurehere}
}

\vspace{-0.5cm}
\section{Discussion}

 The mechanical model presented shows promising results for a telescoping tendon-\textcolor{black}{actuated} robot system when the inner robot is fully sheathed. However, both average and max error increase as the inner robot translates out. This can be due to an inaccurate assumption for the Elastic Modulus, $E$, of the inner robot, where only the $E$ of the nitinol backbone is considered (excluding the inner robot plastic). Additionally, using ArUco tags for image-based position measurement may introduce error. Small inaccuracies, such as tendon misalignment at the base, may also compound the tip position error with increasing robot length. 
 
\textcolor{black}{The current TACTER design only allows for single planar bending. Although this was a clinically driven constraint, it requires manual rotation for a 3D workspace and prevents both robots from bending about different bending planes.}

\textcolor{black}{The cadaver experiments demonstrated that a compliant robot design is necessary to traverse between the middle turbinate and nasal septum to reach the sphenoid. However, the current robot requires more length and an additional DoF to face and reach the sella turcica. This can be done by adding a third bidirectional tube inside the inner tube, and the same mechanical model can be used to predict the tip position of a 3-tube, 5-tendon system.}

\section{Conclusion}
\textcolor{black}{A novel tendon-actuacted concentric tube robot with follow-the-leader capability for EEA surgeries is introduced.} A mechanical model that inputs tendon tension to predict a two-tube robot position is presented and tested in a physical experiment. The average error for six different \textcolor{black}{robot configurations} tested is under 5 mm. However, the max error can be as large as 11.86 mm, which will be reduced in future iterations via closed-loop control and improved model parameters. \textcolor{black}{Cadaver experiments showed the necessity of a 2-DoF system with a compliant inner robot.}

\textcolor{black}{Future work will include adding a third innermost robot, improving the inner robot stiffness parameter, and implementing closed-loop control using electromagnetic (EM) trackers for tip position and FBG sensing for shape estimation. The next iteration of TACTER will also include inner robot rotation and an inner optical fiber for minimally invasive sensing and visualization.}

\section{Acknowledgments}
\textcolor{black}{This work is} supported by the National Science Foundation (NSF) Graduate Research Fellowship Program (GRFP) and the Duke University Traineeship in Advancing Surgical Technologies (TAST) NSF Research Traineeship (NRT). \textcolor{black}{We thank the Duke Surgical Education and Activities Lab (SEAL) for assisting in the cadaver experiments.}

\bibliography{sample}

\noindent\includegraphics[width=1in]{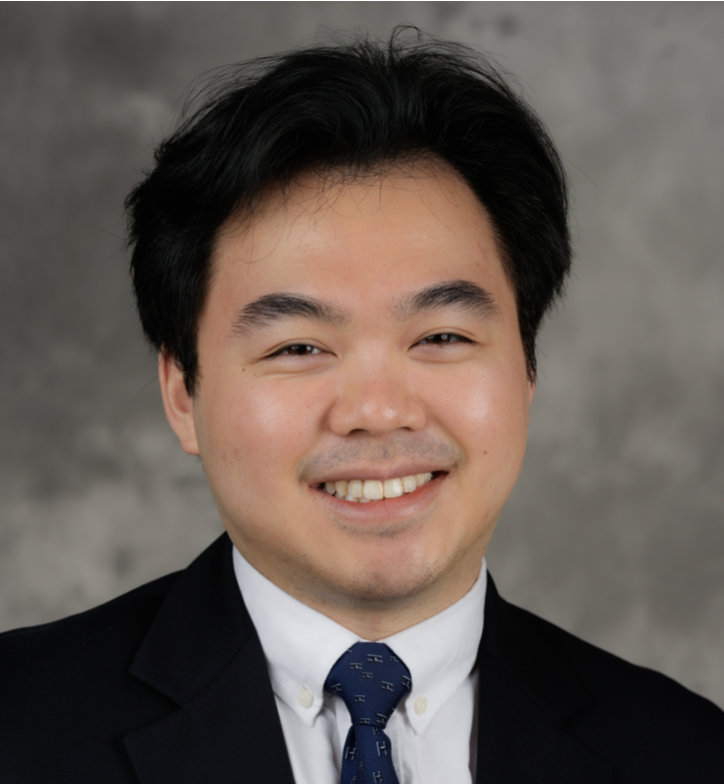}{\bf Kent K. Yamamoto} received his BS in Biomedical Engineering at the Georgia Institute of Technology in 2021. He is currently pursuing his PhD in Mechanical Engineering at Duke University co-advised by Dr. Patrick Codd and Dr. Yash Chitalia from the University of Louisville. His research interests include mechanical design, surgical laser steering, and continuum robotics for minimally invasive surgery.

\noindent\includegraphics[width=1in]{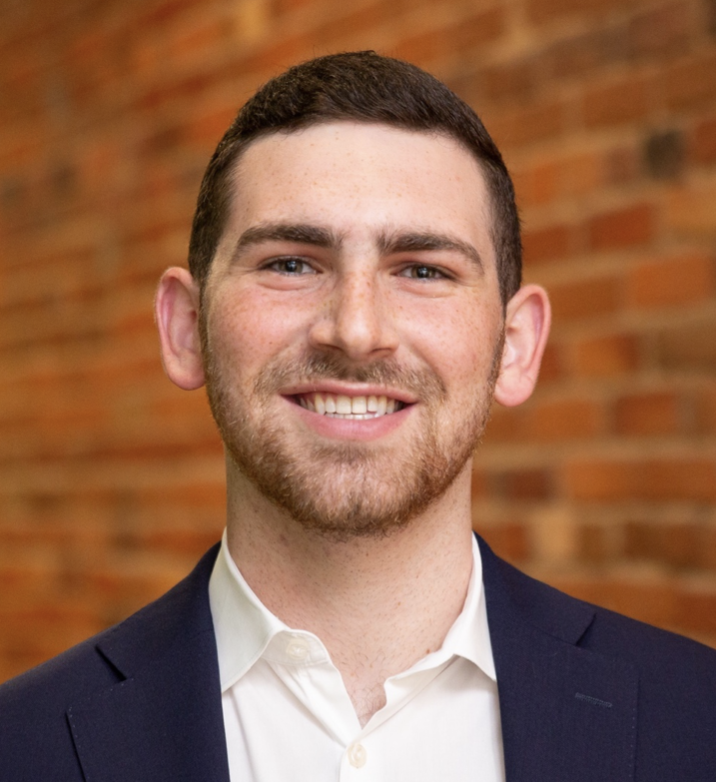}{\bf Tanner J. Zachem} received his BSE in Mechanical Engineering and Materials Science from Duke University. He is currently pursuing a PhD in Mechanical Engineering at Duke University under Drs. Patrick J. Codd and C. Rory Goodwin before completing his medical training. His research lies in developing intraoperative tools, laser systems, machine learning models, and robots to make surgery more data-driven and to increase patient safety. 

\noindent\includegraphics[width=1in]{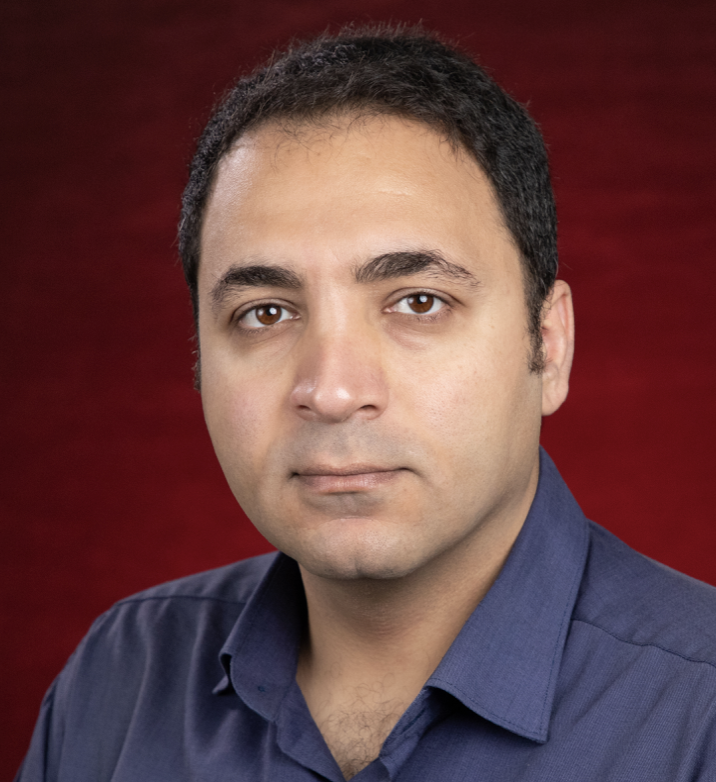}{\bf Pejman Kheradmand} holds a BS in mechanical engineering from Razi University in Kermanshah, Iran, followed by a MS from the University of Tehran in Tehran, Iran. He is currently pursuing a Ph.D. in mechanical engineering at the University of Louisville, Kentucky, USA. His research interests lie in the field of continuum robots in surgical contexts.

\noindent\includegraphics[width=1in]{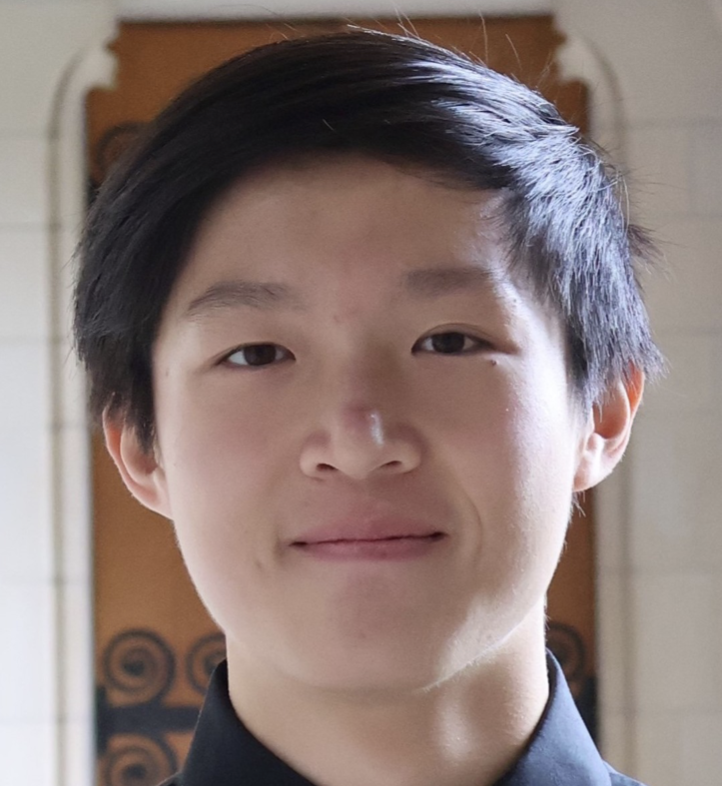}{\bf Patrick Zheng} is currently working towards a BS in Electrical and Computer Engineering and Computer Science with the Pratt School of Engineering, Duke University, Durham, NC, USA. He is currently an undergraduate researcher with the Brain Tool Laboratory. His research interests include surgical robotics, computer vision, and automated robot navigation.

\noindent\includegraphics[width=1in]{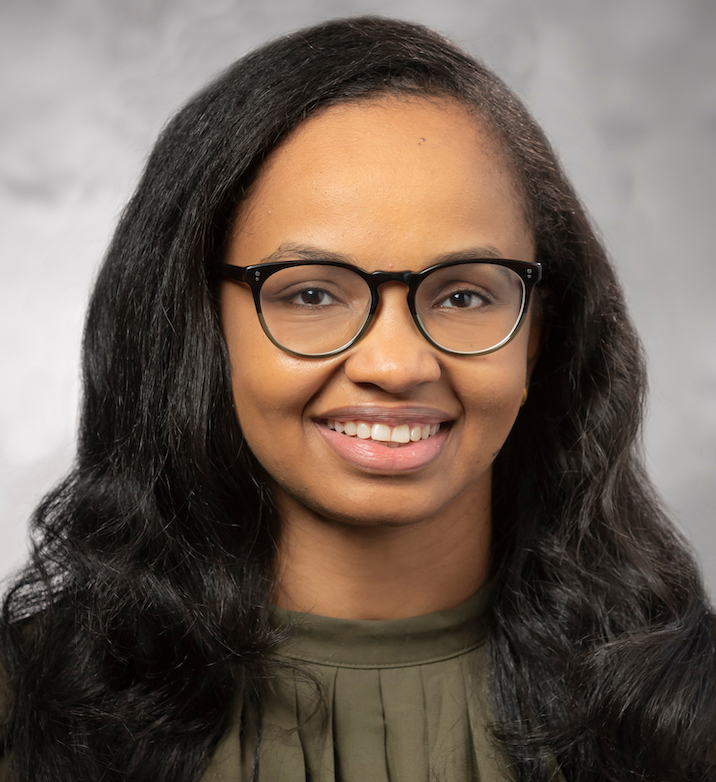}{\bf Jihad Adbelgadir} earned her medical degree from University of Khartoum, and master of science in global health from Duke University. She is currently finishing her neurosurgery training at Duke before pursuing a fellowship in Complex Skull base surgery. Her research interests lie in using computational methods to improve patient care. 

\noindent\includegraphics[width=1in]{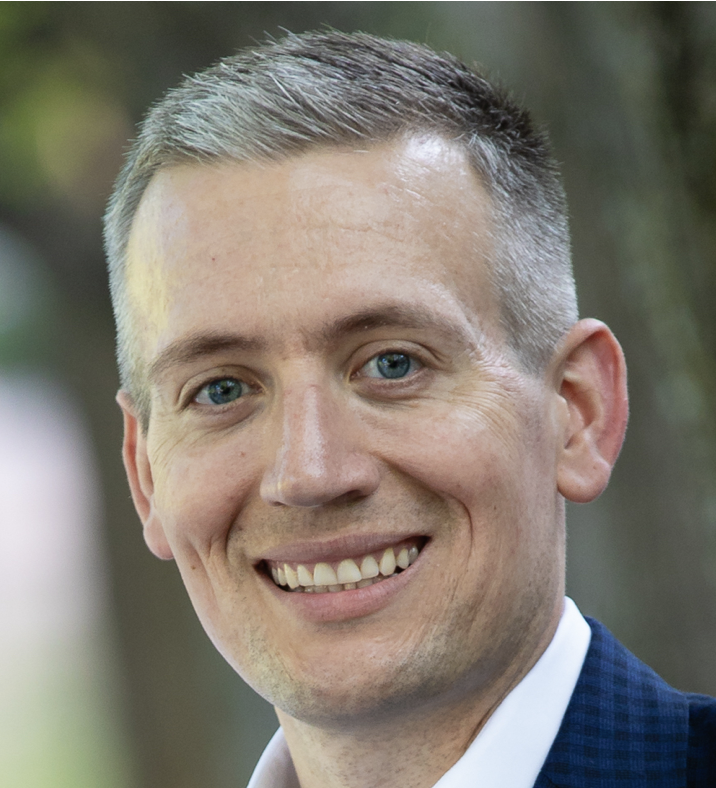}{\bf Jared Laurance Bailey} is pursuing his Master’s in Artificial Intelligence at Duke’s Pratt School of Engineering, as well as a Graduate Certificate in Robotics and Autonomy. Jared's interests include statistical modeling, AI, human-robot interaction, and autonomous robotics.

\noindent\includegraphics[width=1in]{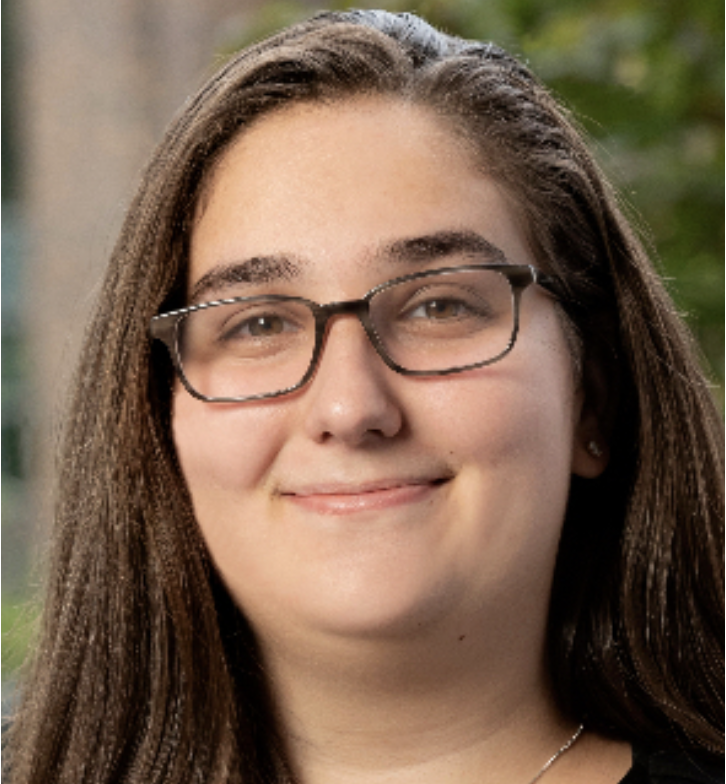}{\bf Kaelyn Pieter} is currently working toward the BS degree in mechanical engineering with the Thomas Lord Department of Mechanical Engineering and Materials Science, Pratt School of Engineering, Duke University, Durham, NC, USA. She is currently a Clark Scholar at Duke. Her research interests include robotics and surgical robotics.

\noindent\includegraphics[width=1in]{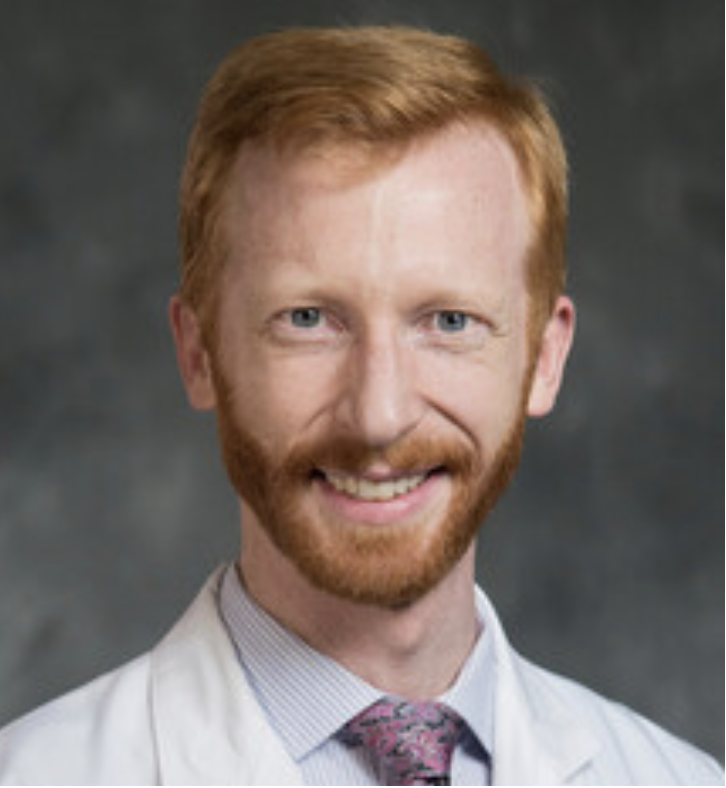}{\bf Patrick J. Codd} received his BS in Biology at the California Institute of Technology and his M.D. with Honors from the Harvard Medical School/Massachusetts Institute of Technology Health Science and Technology Program. He then completed his neurosurgical training at the Massachusetts General Hospital/Harvard Medical School. Dr. Codd is a Board-Certified neurosurgeon and currently is in practice at Duke University Hospital's Preston Robert Tisch Brain Tumor Center. He also serves as an Associate Professor of Mechanical Engineering and Materials Science within the Duke Pratt School of Engineering.

\indent Dr. Codd also served as Chief Medical Officer in two companies (MedXTools Inc.;Mente Inc.).

\noindent\includegraphics[width=1in]{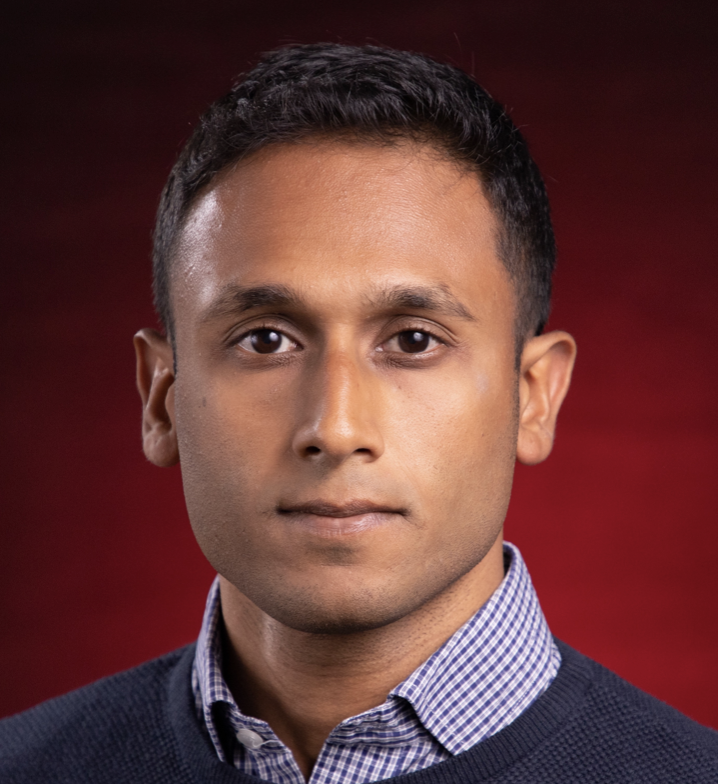}{\bf Yash Chitalia} received his B.S. in Electronics Engineering from the University of Mumbai, Mumbai, India, in 2011, his M.S. degree in electrical engineering
from the University of Michigan, Ann Arbor, MI, USA, in 2013, and a Ph.D. degree in Mechanical Engineering from the Georgia Institute of Technology, Atlanta, GA, USA, in 2021. He is currently an Assistant Professor at the
Department of Mechanical Engineering, University of Louisville, Louisville, KY, USA. His research interests include microscale and mesoscale surgical robots for cardiovascular and neurosurgical applications and rehabilitation
robotics. 
\indent Dr. Chitalia was the recipient of several awards including the Ralph E. Powe Junior Faculty Enhancement Award and the NASA-KY Research
Initiation Award.

\bibliographystyle{ws-jmrr}

\end{multicols}
\end{document}